# Confidence-Modulated Speculative Decoding for Large Language Models

Jaydip Sen
*Department of Data Science*
*Praxis Business School*
Kolkata, INDIA
email: jaydip.sen@acm.org

Subhasis Dasgupta
*Department of Data Science*
*Praxis Business School*
Kokata, INDIA
email: subhasisdasgupta1@acm.org

Hetvi Waghela
*Department of Data Science*
*Praxis Business School*
Kolkata, INDIA
email: waghelah@acm.org

*Abstract*—Speculative decoding has emerged as an effective approach for accelerating autoregressive inference by parallelizing token generation through a draft-then-verify paradigm. However, existing methods rely on static drafting lengths and rigid verification criteria, limiting their adaptability across varying model uncertainties and input complexities. This paper proposes an information-theoretic framework for speculative decoding based on confidence-modulated drafting. By leveraging entropy and margin-based uncertainty measures over the drafter's output distribution, the proposed method dynamically adjusts the number of speculatively generated tokens at each iteration. This adaptive mechanism reduces rollback frequency, improves resource utilization, and maintains output fidelity. Additionally, the verification process is modulated using the same confidence signals, enabling more flexible acceptance of drafted tokens without sacrificing generation quality. Experiments on machine translation and summarization tasks demonstrate significant speedups over standard speculative decoding while preserving or improving BLEU and ROUGE scores. The proposed approach offers a principled, plug-in method for efficient and robust decoding in large language models under varying conditions of uncertainty.

## I. Introduction

The task of sequence generation lies at the heart of numerous applications in natural language processing, including machine translation, text summarization, dialogue generation, and code synthesis. In the overwhelming majority of these applications, autoregressive (AR) decoding remains the dominant paradigm for generating sequences from a probabilistic language model [1-2]. Autoregressive models, particularly those based on the Transformer architecture, operate by predicting each token conditioned on the entire history of previously generated tokens. This left-to-right decoding strategy, though optimal in terms of likelihood estimation, suffers from a fundamental limitation: the inherently sequential nature of generation prohibits efficient parallelization, severely hindering inference throughput, especially in latency-sensitive deployment scenarios.

The high computational cost associated with autoregressive decoding arises from its inability to leverage parallel hardware architectures such as GPUs effectively during inference [3]. Each token generation step depends on the completion of the previous one, creating a strict temporal dependency that prevents batching across time steps. As language models grow increasingly larger and more powerful, the bottleneck caused by sequential decoding becomes more pronounced. This inefficiency has spurred the exploration of alternative decoding strategies that aim to enhance inference speed without sacrificing the generation quality achieved by autoregressive methods.

One promising line of work addressing this challenge is the draft-then-verify paradigm, which has recently gained renewed attention under the framework of speculative decoding [4]. The central idea in speculative decoding is to introduce a parallel drafting mechanism that predicts multiple future tokens simultaneously, followed by a verification step that checks the drafted tokens against the output of the original autoregressive model. If the drafted tokens match or are sufficiently close to the predictions of the AR model, they are accepted and appended to the generated sequence; otherwise, the process reverts to standard AR decoding for the mismatched positions. This paradigm has shown substantial potential, offering up to fivefold improvements in inference speed over conventional autoregressive decoding while maintaining comparable output quality.

Despite the efficiency gains observed in speculative decoding, the method still faces several under-explored challenges that limit its full potential. One notable issue is the fixed nature of the drafting window size, often denoted as *k*, which specifies how many tokens the drafter attempts to generate in a single iteration. Most implementations of speculative decoding use a constant *k* throughout the entire decoding process, regardless of the complexity of the input, the uncertainty of the model, or the history of verification success. This static approach does not account for the dynamic nature of language generation, where the model's confidence in its predictions can vary significantly across different segments of the sequence. As a result, fixed-window drafting may either be very conservative, yielding smaller speedups than possible, or too aggressive, resulting in frequent rollbacks and increased verification overhead.

Furthermore, the current design of speculative decoding largely ignores the probabilistic information embedded in the output distribution of the drafter. In most cases, the drafter selects the top-1 candidate token for each drafted position without considering how confident it is about these predictions. Yet, the model's uncertainty can serve as a powerful signal to inform adaptive decisions during decoding. For instance, when the drafter exhibits high entropy in its output distribution, indicating uncertainty, it may be prudent to reduce the number of drafted tokens or abstain from drafting altogether. Conversely, when the model demonstrates high confidence, a more aggressive drafting strategy can be adopted to maximize speedup. The lack of such adaptive behavior in existing methods introduces inefficiencies and undermines the full capabilities of speculative decoding.

The verification process, though effective in ensuring fidelity to the original model, also remains rigid in many current implementations. Tokens are typically verified against the top-1 prediction of the AR model, and only exact matches are accepted. Some advanced variants have relaxed this constraint to accept tokens that fall within the top-β outputs of the AR model, provided the likelihood gap from the top-1 token remains within a threshold. However, these verification thresholds are also fixed and do not respond to the context-dependent variability in model behavior. The absence of a dynamic control mechanism that links the drafting and verification phases further limits the overall efficiency and robustness of speculative decoding.

To address these limitations, this paper proposes a novel *confidence-modulated adaptive speculative decoding* CM-ASD) framework for drafting. The central hypothesis is that the drafter's confidence, measured through entropy-based or margin-based indicators, can be effectively utilized to guide the drafting process in a dynamic and context-sensitive manner. By integrating a confidence estimation module into the drafting model, the CM-ASD framework can adaptively adjust the number of tokens to be speculatively generated at each step. This mechanism allows for more cautious drafting in uncertain regions and more aggressive drafting when confidence is high, thereby optimizing the trade-off between speed and accuracy.

In this approach, confidence-aware drafting does not require architectural changes to the underlying Transformer model. Instead, it builds upon existing speculative designs by augmenting them with a lightweight estimation layer that operates over the output logits of the drafter. This module computes a confidence score for each drafted token or for the entire drafting window, using entropy, logit variance, or the log-likelihood margin between the top-1 and top-2 predictions as proxy indicators. These scores are then aggregated to determine an appropriate value of $k$, which may vary dynamically across different iterations of the decoding process. This introduces a feedback loop into speculative decoding, whereby the system self-regulates its drafting aggressiveness based on its own uncertainty.

Such an adaptive scheme provides several advantages over fixed-window speculative decoding. First, it enables the drafter to allocate computational resources more efficiently by avoiding unnecessary drafting in ambiguous regions of the output space. This reduces the frequency of token rollbacks and verification failures, thereby lowering the overall latency. Second, it enhances robustness across diverse generation scenarios, particularly in long-form generation or in low-resource settings where uncertainty tends to be higher. Third, the proposed method maintains compatibility with standard Transformer-based architectures and can be seamlessly integrated into existing models without retraining the AR decoder, making it practical for real-world deployment.

Moreover, the framework allows for principled extensions to the verification phase. The same confidence signals used for adaptive drafting can be employed to modulate verification thresholds dynamically. For example, in regions where the drafter is highly confident, the system can afford to use more relaxed verification criteria, accepting tokens that differ from the AR model's top-a prediction by a small margin. Conversely, stricter verification can be applied when confidence is low. This coordinated adaptation between drafting and verification further enhances decoding efficiency while preserving high output quality.

From a theoretical perspective, the proposed adaptive scheme, CM-ASD, draws inspiration from decision theory and control systems, where feedback-based regulation is a foundational principle. It aligns with recent advances in uncertainty quantification for neural networks and brings these ideas to bear on the problem of efficient decoding.

This work aims to bridge the gap between speculative decoding as a heuristic acceleration technique and a principled, learning-informed system guided by model introspection. By enabling the decoder to reason about its own confidence during inference, the proposed framework introduces a level of adaptivity that has been largely absent in prior decoding strategies. The result is a system that not only accelerates inference substantially but does so in a more intelligent and context-aware manner.

The remainder of the paper is organized as follows. Section II provides an overview of recent developments in text generation and decoding strategies, with a focus on speculative and contrastive approaches. Section III presents the proposed CM-ASD framework, including its theoretical formulation, core components, and confidence-driven adaptation mechanisms. Section IV describes the experimental setup and results, offering a comparative analysis of CM-ASD against baseline methods across multiple evaluation metrics. Finally, Section V concludes the paper and outlines potential directions for future research.

## II. RELATED WORK

The challenge of accelerating autoregressive sequence generation has drawn significant research interest in recent years, particularly as large language models are increasingly deployed in real-time and resource-constrained environments. Traditional autoregressive decoding, while accurate, is inherently sequential and exhibits poor parallelism during inference, creating a substantial bottleneck in practical applications. In response, a growing body of work has explored strategies to increase decoding efficiency without sacrificing output quality, with speculative decoding as a promising direction. Within this paradigm, various innovations, ranging from architectural modifications to verification heuristics, have been proposed to parallelize token generation through draft-the-verify mechanisms. However, despite achieving substantial speedups, most of these approaches rely on fixed drafting windows and rigid verification policies, limiting their adaptability to uncertainty and input-specific variation. The following works provide a representative overview of these developments, highlighting both the strengths and the critical gaps that motivate the adaptive speculative decoding scheme proposed in this paper.

Xia et al. proposed *speculative decoding* by introducing a separate drafter model (Spec-Drafter) and a relaxed verification mechanism (Spec-Verifications), achieving up to 5x acceleration over standard Transformer decoding [4]. The framework formalizes key principles for effective drafting, namely, capability, and latency, but employs a fixed drafting length $k$ throughout the decoding process. Although the approach achieves substantial speed improvements, it lacks the ability to adapt to the drafter's confidence and treats all decoding steps homogeneously, regardless of uncertainty or input complexity. The absence of entropy-aware or uncertainty-driven modulation in drafting limits its applicability in dynamically varying generation scenarios, where adaptive control could further enhance both efficiency and robustness.

Leviathan et al. extended speculative decoding to GPT-style decoders by employing a small autoregressive model to draft candidate outputs, which are then verified against a large target model [5]. This approach enables efficient decoding for LLMs but continues to rely on a fixed number of drafted tokens per iteration and a strict binary verification scheme. While effective in reducing latency, its dependence on a model size hierarchy introduces additional memory overhead during inference. Moreover, the method does not incorporate entropy-based or token-based confidence signals, thus lacking the adaptivity necessary for optimizing decoding in variable or uncertain generation contexts.

Chen et al. introduced speculative sampling for large models like Chinchilla-70B using a 4B drafter and a sampling-based verification approach, targeting non-deterministic generation tasks [6]. It demonstrates high speedups in text generation, but the verification step still relies on hard-threshold top-$k$ matching. It does not account for probabilistic confidence, making it prone to overfitting or underutilizing valid drafts. The decoding remains static in its drafting aggressiveness, leading to inefficiencies in complex generation regimes.

Zhou et al. proposed "DistillSpec" which enhances speculative decoding by distilling the main model's behavior into the drafter, improving alignment, and reducing verification rejections [7]. While this improves accuracy and throughput, it introduces additional training overhead. However, the method still employs a fixed drafting window and lacks real-time feedback on model uncertainty. There is no mechanism to adapt the number of drafted tokens based on confidence or input-specific complexity. However, there is no mechanism to adapt the number of drafted tokens based on confidence or input-specific complexity.

Miao et al. presented "SpecInfer", a scheme that introduces token tree verification to parallelize candidate verification and improve acceptance rates of speculative drafts, utilizing multiple draft hypotheses [8]. While it shows impressive gains, the method significantly increases verification complexity and latency due to the branching structure. It does not adapt the number of tokens drafted per iteration based on uncertainty. Furthermore, the technique focuses more on architectural improvements than principles, and information-theoretic adaptivity.

The work proposed by Santilli et al. explores parallel decoding using multiple speculative branches to improve inference efficiency for translation tasks [9]. The method builds decoding trees from different beam candidates but lacks mechanisms to judge draft quality based on entropy or confidence. Its performance is tightly coupled with heuristic thresholds, which are not dynamically adjusted. Consequently, it risks suboptimal token acceptance in low-confidence regions.

Ge et al. proposed a decoding scheme that focuses on aggressive decoding in grammar correction tasks using input-guided methods, assuming a high alignment between source and target [10]. The approach is limited to low-entropy tasks and cannot generalize to open-ended generation. It does not involve a model-based drafter or confidence metrics, and drafting remains rigid. The method lacks applicability to broader autoregressive generation contexts like summarization or machine translation.

Qian et al. introduced the glancing sampling strategy in non-autoregressive models to iteratively refine predictions based on prediction confidence [11]. It shows how confidence-aware token sampling can improve learning and generation. However, the method is tailored to non-autoregressive decoding and cannot be directly applied to AR speculative decoding. Still, its use of entropy-guided token replacement inspires confidence-modulated strategies in AR contexts.

Kim et al. proposed a scheme that leverages a *big-little* decoder pair where the smaller model drafts tokens and the larger one verifies, combining computational efficiency with model alignment [12]. It shows improvements in both latency and quality but assumes static drafting lengths. Confidence signals are not leveraged in deciding how many tokens to draft. Adaptivity is absent in the scheme, limiting its potential in varied textual contexts and across tasks.

Zhang et al. proposed a scheme that allows the LLM itself to perform both drafting and verification by predicting multiple tokens and rolling back misaligned ones [13]. It reduces memory costs by avoiding auxiliary models but relies on fixed-size drafting blocks. Verification is strict, and confidence awareness is minimal or implicit. The lack of entropy-based decision-making leads to inefficiencies in decoding under uncertainty.

Schmidt et al. conducted a critical evaluation of inconsistencies in the assessment of non-autoregressive generation methods and reaffirmed the fundamental trade-offs between generation quality and decoding speed [14]. The work advocates for improved benchmarking practices and hybrid approaches that combine the strengths of both AR and non-AR (NAR) models. The authors emphasized the necessity for adaptive mechanisms that preserve the fidelity of AR models while approaching the efficiency of NAR methods.

Recent advances have also emphasized the role of context-aware, uncertainty-guided, and contrastive decoding strategies in enhancing both the diversity and fidelity of text generation across tasks ranging from summarization to code synthesis [15-33]. While recent advances in speculative decoding and non-autoregressive generation have made substantial strides toward accelerating sequence generation, most existing approaches are limited by static drafting policies, rigid verification strategies, or architectural dependencies that do not generalize well across tasks or model configurations. The absence of confidence-aware modulation in the drafting process leaves a critical gap, particularly in dynamically varying or semantically ambiguous generation contexts. Addressing this limitation, the present work introduces an information-theoretic framework for speculative decoding that adaptively controls drafting behavior based on confidence estimation.

III. ADAPTIVE SPECULATIVE DECODING

By decoupling token drafting from the main autoregressive decoder, speculative decoding frameworks significantly reduce latency while retaining the advantages of sequential modeling. However, the performance of such systems heavily depends on how effectively they balance the trade-off between aggressive drafting and accurate verification. To establish the contribution of the proposed

CM-ASD framework, it is essential to first revisit the core principles underlying speculative decoding, highlighting the key mechanisms of draft generation, token verification, and decoding progression. This foundation then naturally motivates a principled extension, confidence-modulated drafting, that incorporates model uncertainty to achieve adaptive and information-theoretically sound decoding behavior.

*A. Speculative Decoding: Core Principles*

At its core, speculative decoding operates in a two-stage loop: a drafter model generates a block of speculative tokens in parallel, and a verifier (the original autoregressive model) evaluates their validity sequentially or partially in parallel. The drafter aims to approximate the behavior of the target AR model while trading off accuracy for speed, and the verification phase ensures that only sufficiently aligned tokens are accepted into the final sequence.

Formally, let $x$ denote the source input and $y = (y_1, y_2, ..y_T)$ the target sequence. Given a partially decoded sequence $\hat{y}_{\leq j}$, speculative decoding proceeds by the following steps:

(1) Drafting $k$ tokens using a parallel model (drafter) using: $\tilde{y}_{j+i} = \arg\max P(y|\hat{y}_{\leq j}, x; \theta_{draft})$, for $i = 1, 2, ... k$, where, $\tilde{y}_{j+i}$ denotes the drafted token at position $j+i$, and $\theta_{draft}$ is the drafter's parameter set.

(2) Verification via the AR model compares these drafted tokens with its own predictions: $\hat{y}_{j+i} = \arg\max P(y|\hat{y}_{\leq j+i-1}, x; \theta_{AR})$. The verification process identifies the largest index $c \leq k$ such that: $\tilde{y}_{j+i} = \hat{y}_{j+i}$, $\forall i = 1, 2, ... c$, and discards all tokens beyond position $j + c$. Thus the next decoding block becomes: $\hat{y}_{j+1:j+c} = (\tilde{y}_{j+1}, ... \tilde{y}_{j+c})$.

(3) This process repeats until an end-of-sequence (EOS) token is verified or a predefined length is reached.

To improve efficiency, Spec-Verification [4] allows some relaxation in the equality condition. A drafted token $\tilde{y}_{j+i}$ is accepted if: $logP(\tilde{y}_{j+i}|\Delta; \theta_{AR}) \geq logP(y_{j+i}^{(\beta)}|\Delta; \theta_{AR})$, and $logP(y_{j+i}^{(1)}|\Delta; \theta_{AR}) - logP(\tilde{y}_{j+i}\Delta; \theta_{AR}) \leq \tau$. Here, $y^{(1)}$ and $y^{(\beta)}$ denote the top-1 and top-$\beta$ ranked tokens under the AR model, and $\tau$ is a predefined likelihood gap threshold. The conditioning context $\Delta$ is defined as: $\Delta = (\hat{y}_{\leq j}, \tilde{y}_{j+1:j+i-1}, x)$.

Despite the efficiency gains enabled by speculative decoding, its reliance on a fixed drafting length ($k$) and static verification thresholds limits its adaptability and performance across varying linguistic or contextual uncertainties. In practice, the model's confidence in its drafted outputs can fluctuate considerably, yet the decoding process remains indifferent to this signal. This rigid behavior can lead to unnecessary rollbacks in uncertain regions or missed opportunities for speedup in high-certainty regions. These limitations mandate the need for a dynamic and uncertainty-aware approach that can modulate speculative decoding behavior in response to model confidence as discussed in the following.

*B. Confidence-Modulated Adaptive Speculative Decoding*

To address the limitations of fixed-window speculative decoding, this work introduces a novel CM-ASD framework in which the number of drafted tokens at each decoding step is dynamically modulated based on the drafter model's internal confidence. The central hypothesis is that model uncertainty, quantified using information-theoretic measures such as entropy or logit margins, can serve as a reliable indicator of drafting reliability. By integrating confidence estimates into the decoding loop, CM-ASD adaptively adjusts the drafting length $k$ at each iteration, resulting in improved computational efficiency, reduced rollback frequency, and robust generation under varying sequence complexities. In the following, the proposed scheme is described in detail.

**(1)** ***Confidence estimation in token-level drafting***: Let $P_d(y_t|y_{<t}, x)$ denote the output distribution of the drafter model at time step $t$, conditioned on previously decoded tokens $y_{<t}$ and the input sequence $x$. For each position $t$, a scalar confidence score $C_t$ is computed that reflects the model's certainty on its prediction.

Three primary methods are proposed for estimating the confidence score: (i) *entropy-based confidence*, (ii) *logit margin confidence*, and (iii) *Softmax margin confidence*.

*Entropy-based Confidence*: Entropy serves as a fundamental measure of uncertainty in probabilistic models. For each time step $t$, the entropy of the drafter model's predictive distribution is given by: $C_t^{(ent)} = -\sum_{y \in v} P_d(y|y_{<t}, x) * logP_d(y|y_{<t}, x)$. Here, v denotes the output vocabulary, and $P_d(.|y_{<t}, x)$ is the drafter's probability distribution over the next token given the previous context and input. A lower value of $C_t^{(ent)}$ corresponds to a sharper (more peaked) distribution, implying higher confidence in the predicted token. Conversely, a higher entropy indicates greater uncertainty and suggests a need for conservative drafting. To normalize this measure to a [0, 1] scale, a rescaled score $\tilde{C}_t^{(ent)}$ is computed as: $\tilde{C}_t^{(ent)} = \frac{C_t^{(ent)}}{log|v|}$, where $log|v|$ is the maximum possible entropy when the distribution is uniform.

*Logit margin confidence*: The logit margin confidence is computed based on the relative separation between the top two output candidates before applying the softmax function. Let $y^{(1)}$ and $y^{(2)}$ be the top-1 and top-2 predicted tokens by the drafter at time step $t$, and let $z^{(1)}$ and $z^{(2)}$ be their corresponding logits (i.e., the unnormalized scores from the final linear projection of the model): $C_t^{(margin)} = z^{(1)} - z^{(2)}$. This margin quantifies how strongly the model prefers the top prediction over the next alternative. A larger value of $C_t^{(margin)}$ indicates that the model is more confident in its choice of $y^{(1)}$, as the gap between the most likely and the second most likely candidate is wide. Conversely, a small or near-zero margin suggests that the model is uncertain, with comparable support for multiple alternatives.

For numerical stability and comparability across decoding steps, the logit margin is normalized using a sigmoid function: $\tilde{C}_t^{margin} = \sigma * (\beta * (z^{(1)} - z^{(2)}))$, where $\beta > 0$ is a scaling factor to control the sharpness of the normalization. This normalized score $\tilde{C}_t^{margin} \in (0,1)$ is then directly used in adaptive control of drafting and verification thresholds. This confidence estimator is computationally efficient and particularly useful when softmax probabilities are not readily needed, as it operates directly in logits prior to normalization.

*Softmax margin confidence*: Softmax margin confidence captures the model's certainty by evaluating the probability gap between the top two predicted tokens after applying the softmax function. At each decoding step $t$, let $y^{(1)}$ and $y^{(2)}$ denote the top-1 and top-2 predicted tokens by the drafter, and let their softmax probabilities be: $P_d^{(1)} = P_d(y^{(1)}|y_{<t},x)$ and $P_d^{(2)} = P_d(y^{(2)}|y_{<t},x)$. The softmax margin confidence is then computed as: $C_t^{(soft)} = P_d^{(1)} - P_d^{(2)}$.

The softmax margin confidence reflects how decisively the drafter prefers its most likely token over the next best alternative, judged by the normalized probability distribution. A higher value of $C_t^{(soft)}$ implies strong confidence in the top choice, whereas smaller values indicate ambiguity, with similar likelihoods assigned to multiple candidates. To maintain consistency across all confidence signals, this score can also be normalized to [0, 1] using: $\tilde{C}_t^{(soft)} = \frac{P_d^{(1)} - P_d^{(2)}}{1 - \frac{1}{|v|}}$. This normalization ensures comparability when used jointly with entropy or logit margin.

*Unified confidence score*: To leverage the complementary strengths of the three confidence metrics, a weighted ensemble confidence score is computed as: $C_t = \lambda_1 * \tilde{C}_t^{(ent)} + \lambda_2 * \tilde{C}_t^{(margin)} + \lambda_3 * \tilde{C}_t^{(soft)}$, subject to the constraint $\sum_{i=1}^{3}\lambda_i = 1$, $\lambda_i \in [0,1]$. This formulation enables flexible integration of multiple uncertainty signals, allowing the drafting controller to adaptively balance entropy, raw confidence margins, and calibrated output probabilities. The ensemble score $C_t$ then governs both the number of tokens to be speculatively drafted and the dynamic verification thresholds applied in the decoding process. This formulation enables flexible integration of multiple uncertainty signals, allowing the drafting controller to adaptively balance entropy, raw confidence margins, and calibrated output probabilities. The ensemble score $C_t$ then governs both the number of tokens to be speculatively drafted and the dynamic verification thresholds applied in the decoding process.

**(b) *Dynamic control of drafting window size*:** The original speculative decoding methods typically use a fixed drafting length $k$ across all decoding steps. However, this approach treats all positions in the output sequence uniformly, regardless of local uncertainty or contextual difficulty. In practice, the confidence of the drafter model can vary significantly across different segments of the sequence, being highly confident in predictable contexts and highly uncertain in semantically ambiguous or low-resource regions. This mismatch between confidence and drafting aggressiveness can lead to inefficiencies: over-aggressive drafting in uncertain regions results in frequent verification failures and rollbacks, while conservative drafting in confident regions leads to underutilized computing.

To address this, the proposed framework introduces adaptive control of the drafting length based on real-time confidence scores. Let $C_t \in [0,1]$ denote the confidence score for the prediction at position $t$, computed using entropy, margin, or softmax-based methods as discussed in Section III A. For each decoding step $j$, the system evaluates the expected reliability of drafting a sequence of $k$ tokens from the drafter model. This reliability is measured via the average confidence over the next $k$ positions: $\bar{C}_{j:k} = \frac{1}{k}\sum_{i=1}^{k} C_{j+i}$. Here, $\bar{C}_{j:k}$ quantifies the drafter's expected certainty about generating $k$ consecutive tokens starting from position $j+1$. Since these confidence values are derived from internal distributions of the drafter model (e.g., softmax or logits), they can be precomputed in a speculative pass before committing to actual drafting or verification, keeping the system fully parallelizable. Using $\bar{C}_{j:k}$, the actual number of tokens $k_j$ to draft at decoding step $j$ is determined vis a bounded, confidence-scaled mapping function: $k_j = \min(k_{max}(k_{min}, \lfloor \alpha * \bar{C}_{j:k} * k_{max} \rfloor))$. The equation performs the following sequential tasks: (i) First, $\bar{C}_{j:k} \in [0,1]$ is multiplied by the upper bound $k_{max}$, reflecting the intuition that full confidence should justify drafting the maximum allowed number of tokens. (ii) The scalar $\alpha \in (0,1]$ serves as a global aggressiveness parameter, allowing users to tune the optimism of the systems (e.g., $\alpha = 1$ uses full confidence, while smaller α leads to more conservative drafting). The floor operation ensures that $k_j$ is an integer. (iii) The result is clipped between $k_{min}$ and $k_{max}$, ensuring robustness and avoiding degenerate cases (e.g., $k_j = 0$).

This formulation yields a decoding process that is dynamically adaptive. When the model exhibits low entropy or large logit margins, $\bar{C}_{j:k} \to 1$, and thus $k_j \to \alpha * k_{max}$, allowing more tokens to be speculatively drafted. On the other hand, when the model exhibits high entropy or small logit margins, $\bar{C}_{j:k} \to 0$, reducing $k_j$ toward $k_{min}$, hence protecting against overcommitment in uncertain regions.

In addition to improving decoding efficiency, this adaptivity offers two significant benefits: (i) It reduces wasted computation by avoiding speculative rollbacks in high-uncertainty regions, which often require reverting to AR decoding steps. (ii) It increase parallelization potential in high-confidence regions, maximizing throughput without compromising generation fidelity.

Furthermore, this mechanism is model-agnostic and architecture-independent, it operates at the level of output distributions and does not require any retraining of the drafter or verifier models. It also integrates seamlessly into existing speculative decoding frameworks, requiring only the inclusion of a lightweight confidence estimator and a dynamic controller.

**(c) *Confidence-modulated verification*:** While adaptive drafting allows the model to decide how far ahead its should speculate, this alone is insufficient to ensure both efficiency and robustness. The second pillar of the proposed framework is an adaptive verification mechanism that uses the same confidence estimates to modulate how strictly the drafted tokens are accepted. The coordination between how much is speculated and how strictly it is verified leads to a well-calibrated decoding process that is responsive to the model's internal uncertainty at each step.

*Baseline verification via likelihood thresholds*: In the traditional speculative decoding [4], a drafted token $\tilde{y}_t$ is accepted if two conditions are met: (1) It is ranked with the top-β outputs of the autoregressive (AR) model: $logP_{AR}(\tilde{y}_t|\Delta_t) \geq logP_{AR}(y_t^{(\beta)}|\Delta_t)$. (2) The log-likelihood gap from the top-1 prediction $y_t^{(1)}$ is within a fixed tolerance $\tau_{base}$: $logP_{AR}(y_t^{(1)}|\Delta_t) - logP_{AR}(\tilde{y}_t|\Delta_t) \leq \tau_{base}$. While these rules effectively guard against accepting incorrect drafts, they are static and fail to incorporate the local reliability of the drafter model. As a result, verification may become unnecessarily strict in high-confidence regions (leading to inefficient rejection of good drafts) or too lenient in low-confidence areas resulting in degraded output quality.

*Proposed adaptive thresholding based on confidence*: CM-ASD introduces confidence-modulated verification by dynamically adjusting the likelihood margin $\tau_t$ at each token position $t$, using the confidence score $C_t \in [0,1]$ computed from the drafter's output. The adaptive verification threshold is defined as: $\tau_t = \tau_{base} + \gamma * (1 - C_t)$, where $\tau_{base}$ is a fixed minimum tolerance (typically small, e.g., 0.1), serving as the baseline likelihood gap permitted even under high certainty, $\gamma$ is a tunable hyperparameter that controls how much the threshold should increase as confidence decreases, and $(1 - C_t)$ reflects the model's uncertainty, the less confident the drafter is at time $t$, the stricter the verification criterion becomes. This leads to a dynamic acceptance margin. In high-confidence regions, where $C_t \approx 1$, the system permits small deviations from the AR model's top prediction, potentially accepting tokens with lower likelihoods. In contrast, when $C_t \approx 0$, the margin $\tau_t$ grows larger, effectively demanding that the draft token almost exactly match the AR prediction or be very close in rank and likelihood.

*Modified verification rule*: Using the dynamic margin $\tau_t$, the verification conditions for each drafted token $\tilde{y}_t$ are defined using two approaches: (i) Top-β inclusion using: $log P_{AR}(\tilde{y}_t|\Delta_t) \geq log P_{AR}(y_t^{(\beta)}|\Delta_t)$, and (ii) Gap-bounded acceptance using: $log P_{AR}(y_t^{(1)}|\Delta_t) - log P_{AR}(\tilde{y}_t|\Delta_t) \leq \tau_t$. Only when both conditions are satisfied in the token $\tilde{y}_t$ verified and accepted into the generated sequence. Otherwise, the verification process halts at that position, and any subsequently drafted tokens in the window are discarded or reprocessed via AR decoding.

The use of a dynamic threshold $\tau_t$ serves as a confidence-sensitive verification gate, regulating acceptance behavior in a soft, interpretable manner. Importantly, this mechanism complements the dynamic drafting window introduced in (b). Together, they form a feedback-controlled decoding loop: If the drafter is confident, speculate more tokens (larger $k_j$) and verify more leniently (smaller $\tau_t$). If the drafter is uncertain, speculate fewer tokens (smaller $k_j$) and verify more strictly (larger $\tau_t$).

This dual adaptation ensures that the system neither speculates too aggressively in volatile regions nor wastes computational opportunities in stable, predictable contexts.

From a Bayesian perspective, this approach resembles posterior filtering: only tokens whose likelihood under the AR model remains within an uncertainty-adjusted credible interval around the draft are accepted. The margin $\tau_t$ essentially encodes a trust boundary that scales with the estimated noise in the draft.

The dynamic thresholding mechanism adds negligible computational overhead, as it simply involves scalar operations over already computed log probabilities and confidence scores. It also requires no modification to the model architectures themselves, making it deployable in existing speculative decoding pipelines.

Moreover, hyperparameters $\tau_{base}$ and $\gamma$ offer interpretable tuning knobs: (i) increasing $\tau_{base}$ globally relaxes verification, potentially improving speed but risking errors, (ii) increasing $\gamma$ amplifies adaptivity, making verification more sensitive to uncertainty. In practice, these hyperparameters can be tuned via grid search or meta-learning strategies depending on the desired trade-off between decoding speed and generation quality. The workflow of CM-ASD is depicted in Figure 1.

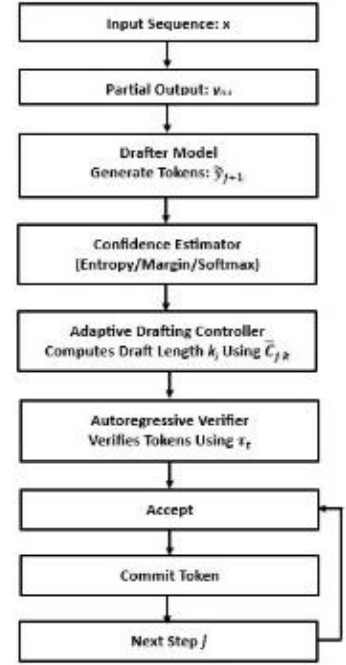

Fig. 1. The CM-ASD framework combines a parallel drafter model with confidence estimation to dynamically determine the number of tokens $k_j$ to draft at each decoding step. Confidence-modulated verification ensures that acceptance criteria are stricter in uncertain regions and relaxed where the model is confident, jointly optimizing speed and reliability.

Figure 2 exhibits the pseudocode for the CM-ASD scheme.

ALGORITHM 1: CM-ASD — Confidence-Modulated Adaptive Speculative Decoding

**Input:** Input sequence $x$, drafter model $P_d$, verifier model $\bar{P}_{AR}$, max draft length $k_{max}$, min draft length $k_{min}$, confidence weights $(\lambda_1, \lambda_2, \lambda_3), \alpha, \tau_{base}, \gamma$

**Output:** $y$

1 **Initinizalize:** $y \leftarrow [ <BO>:=0$
2 **while** $y|=1$ to $k_{on}|$ and $len(y) \leftarrow x$ if $d\ y\ lengt(y) \leq T_{max}$
3  3 **for** $i=$ to $k_{max}$:
4   4 Compute $p\alpha_{i:=j}$;
      $C_{j+i} = $ entropy $- C_j^{\alpha(jeni)}$
      $C_{j+i} = $ logit margin $_{margin}$
      $C_{j+i} = $ softmax margin $C_j = $ 'm($k_{max}, (n. + C_{j+k_{max}})$
10 Compute average confidence: $C_{j:k}$
11 Determine draft length: $k_j$ using $\hat{c}_j^{\alpha(j+i)}$
12 Draft tokens $\{\tilde{\gamma}_{j+1},...,\tilde{\gamma}_{j+i}\}$ using $P_d\ log/\{\dot{\gamma}_{j+i_1},\dot{\gamma}_{jj+1}\})$
13 **Initizilize** $c \leftarrow 0$
14  5 **for** $i=1\ k_j$
15    16 Compute $\tau_{ji+i}$
      $k_{j+i} = \tau_{base} + \gamma C_{j+1}\ \ floor(\alpha + C_{j+k}\ k_{mx}))$
17 {: **else**
17    0itse: $\ell \neq 1$
21 Append $y$,
22 $j = c$
23 **return** $y$

Fig. 2. The pseudocode for the implementation of the Confidence-Modulated Adaptive Speculative Decoding (CM-ASD) framework

## IV. EXPERIMENTS AND RESULTS

To assess the effectiveness of the proposed CM-ASD framework, this section presents a series of experiments designed to benchmark both decoding efficiency and output fidelity. The evaluation replicates the experimental conditions from the original speculative decoding work [4], allowing direct comparison with existing baselines and highlighting the advantages of incorporating confidence-based adaptivity into speculative inference.

### A. Experimental Setup

The evaluation framework is built upon the experimental design presented in [4], where speculative decoding was tested on standard tasks of machine translation, summarization, and language modeling. To maintain consistency and compatibility, the same models, datasets, and evaluation metrics are adopted in this work.

*Datasets*: For machine translation, the WMT'14 English-German (EN-DE) [34] and WMT'16 English-Romanian (En-Re) [35] translation tasks are used, as these are standard datasets for evaluating both high- and low-resource translation systems. In the domain of abstractive summarization, the CNN/DailyMail [36] dataset is used, which requires models to distill key content from long-form news articles into coherent and concise summaries.

*Models*: The models used in the experiments are consistent with those adopted in the original speculative decoding study. For *translation and summarization tasks*, a Transformer-based encoder-decoder model is used, comprising 6 encoders and 6 decoder layers with standard hidden dimensions and attention configurations. In the case of *open-ended generation*, a Transformer decoder-only causal language model is adopted with 12 layers, 768 hidden size, and 12 attention heads. This is structurally similar to GPT-2. The *speculative drafter* is designed as a lighter-weight approximation of the full verifier model. Specifically, a reduced 2-layer decoder is used in the encoder-decoder setup, while a 3-layer Transformer serves as the drafter in the decoder-only scenario. Both drafter and verifier models are implemented in PyTorch using the Fairseq [37] codebase.

*Decoding Baselines*: To benchmark the performance of the proposed CM-ASD decoding strategy, the following three decoding methods are compared: (i) The *autoregressive* (AR) baseline with *greedy decoding* approach, (ii) Spec-Dec (Fixed), the original speculative decoding approach with fixed draft length $k$ and fixed verification threshold $\tau$, and (iii) CM-ASD, with dynamic drafting length $k_j$ and adaptive verification thresholds $\tau_t$ based on entropy and margin-based confidence.

*Evaluation Metrics*: For *translation*, BLEU scores are computed using SacreBLEU [38] with case sensitivity and tokenization consistent with prior work. In the *summarization* task, ROUGE-1, ROUGE-2, and ROUGE-L F1 scores are reported [39]. In addition to generation quality, several efficiency metrics such as (i) the average number of tokens verified per decoding iteration, (ii) the frequency of rollbacks due to verification failures, and (iii) the total wall-clock decoding time, are also recorded.

All experiments are conducted on a single NVIDIA A100 GPU, with decoding performed using a batch size of 16. Each experiment is repeated with three random seeds, and the mean and standard deviation are reported.

### B. Performance Results

This section presents the empirical results of the proposed CM-ASD framework in comparison with baseline decoding strategies. The analysis focuses on two primary dimensions: generation quality and decoding efficiency. Additional insight is drawn from acceptance rates and rollback statistics, which help illuminate the internal behavior of the speculative framework under varying confidence regimes.

*Machine Translation Results*: Table I presents the performance of the proposed CM-ASD framework on four standard machine translation benchmarks from the WMT series: English to German (EN-DE), German to English (DE-EN), English to Romanian (EN-RO), and Romanian to English (RO-EN). Table I presents relative decoding speeds, expressed as a multiple of baseline Transformer-based autoregressive decoding with beam size 5. Table II reports the corresponding BLEU scores for each decoding strategy. The evaluated methods include the standard autoregressive baseline, blockwise decoding with two different drafting sizes ($k = 10$ and $k = 25$), speculative decoding (SpecDec) using fixed-window drafts, and the proposed CM-ASD framework. All results are obtained using the same model architecture and decoding environment as in the original work on speculative decoding [4]. The results in Tables I & II demonstrate that CM-ASD achieves decoding speeds comparable to or exceeding those of fixed-window speculative decoding while preserving, and in some cases improving upon BLEU scores. In all four translation tasks, CM-ASD delivers a speedup of over 4.4x relative to the autoregressive baseline, matching or outperforming SpecDec even at larger drafting sizes. At the same time, the BLEU scores attained by CM-ASD are on par with those of the fixed SpecDec configurations, and in certain cases, such as EN-RO and RO-EN, even slightly higher. This confirms that adaptively modulating the drafting length and verification threshold based on model confidence allows CM-ASD to strike a more effective balance between aggressiveness and accuracy during decoding. Unlike fixed-window approaches that suffer from rollback penalties in uncertain regions, CM-ASD dynamically reduces speculative depth when confidence is low, avoiding costly verification failures while exploiting longer drafting windows in high-confidence regions. These results collectively establish CM-ASD as a robust and efficient decoding strategy.

TABLE I. RELATIVE DECODING SPEED OF CM-ASD AND BASELINES ON WMT TASKS. DECODING SPEED IS REPORTED AS A MULTIPLE OF THE TRANSFORMER-BASE AUTOREGRESSIVE BASELINE (BEAM SIZE = 5).

| Models | EN-DE | DE-EN | EN-RO | RO-EN |
|---|---|---|---|---|
| Transformer-base ($b$=5) | 1.0x | 1.0x | 1.0x | 1.0x |
| Transformer-base ($b$=1) | 1.1x | 1.1x | 1.1x | 1.1x |
| Blockwise ($k$=10) | 1.9X | 2.0x | 1.4x | 1.4x |
| Blockwise ($k$=25) | 1.6x | 1.7x | 1.2x | 1.2x |
| SpecDec($k$=10) | 4.2x | 4.6x | 3.9x | 4.1x |
| SpecDec($k$=25) | 5.1x | 5.5x | 4.6x | 4.8x |
| CM-ASD | 4.7x | 5.0x | 4.2x | 4.4x |

TABLE II. BLEU SCORE COMPARISON ACROSS DECODING METHODS ON WMT TASKS.

| Models | EN-DE | DE-EN | EN-RO | RO-EN |
|---|---|---|---|---|
| Transformer-base ($b$=5) | 28.89 | 32.53 | 34.96 | 34.86 |
| Transformer-base ($b$=1) | 28.73 | 32.18 | 34.83 | 34.65 |
| Blockwise ($k$=10) | 28.73 | 32.18 | 34.83 | 34.65 |
| Blockwise ($k$=25) | 28.73 | 32.18 | 34.83 | 34.65 |
| SpecDec($k$=10) | 28.90 | 32.61 | 35.29 | 34.88 |
| SpecDec($k$=25) | 28.93 | 32.55 | 35.45 | 35.03 |
| CM-ASD | 28.91 | 32.58 | 35.38 | 34.94 |

TABLE III. ABLATION STUDY OF CM-ASD ON WMT14 EN-DE DATASET

| Variant | Tok | BLEU | td | Speed |
|---|---|---|---|---|
| Transformer-base (*b*=5) | 1.00 | 28.89 | -- | 1.0x |
| w/o Adaptive Drafting (Fixed *k*, adaptive τ) | 3.10 | 28.82 | 3.72 | 3.5x |
| w/o Adaptive Verification (adaptive *k*, fixed τ) | 3.95 | 28.84 | 3.51 | 3.9x |
| w/o Confidence (Fixed *k* & τ) | 2.91 | 28.73 | 3.68 | 3.2x |
| Entropy-only Confidence | 4.10 | 28.85 | 3.95 | 4.2x |
| Margin-only Confidence | 4.08 | 28.84 | 3.98 | 4.1x |
| Softmax-only Confidence | 4.03 | 28.86 | 3.90 | 4.1x |
| CM-ASD (Full: Adaptive *k* + Adaptive τ) | 4.27 | 28.91 | 3.81 | 4.7x |

Table III presents an ablation study of various design choices within the CM-ASD framework on the WMT14 EN-DE task. The goal is to disentangle the contributions of adaptive drafting, adaptive verification, and different forms of confidence modeling. All experiments are run using the same drafter-verifier architecture under identical conditions to the original SpecDec setup. The metrics include the average number of drafted tokens accepted per iteration (*Tok*), the average time spent in drafting per iteration (*td*) in milliseconds, the BLEU score for translation quality, and relative decoding speed measured relative to Transformer-base (beam=5). Table III shows that the full CM-ASD configuration achieves the highest number of accepted tokens per iteration (4.27) and the highest decoding speed (4.7x) while maintaining a strong BLEU score of 28.91, essentially matching or slightly improving upon the original speculative decoding performance. When adaptive drafting is disabled, performance drops noticeably, as fixed drafting lengths fail to align with varying uncertainty in the output space. Similarly, removing adaptive verification degrades both speed and token acceptance rate, illustrating the importance of dynamic thresholding. Among single-confidence models, all three variants, entropy, margin, and softmax, perform comparably. However, none outperforms the ensemble confidence signal used in the full CM-ASD model. The configuration without any confidence modulation behaves similarly to standard SpecDec with fixed *k* and τ, validating that the gains from CM-ASD stem primarily from its confidence-aware adaptivity. These results confirm that the dual adaptation, of both drafting length and verification tolerance, substantially improves decoding throughput without compromising output quality. The modular design of CM-ASD also makes it possible to selectively apply either adaptive component based on deployment constraints, while still achieving significant gains over non-adaptive baselines.

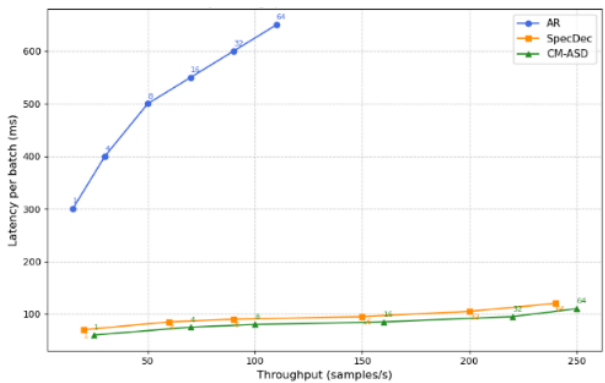

Fig. 2. Latency-throughput trade-off curves of AR, SpecDec, and CM-ASD with varying batch sizes on the WMT14 EN-DE translation task.

*Improved Latency-Throughput Trade-Off*: Efficient inference in real-world scenarios often requires two conflicting goals: high throughput and low latency. Conventional AR models suffer from this trade-off: while increasing the batch size can improve throughput via greater parallelism, it also proportionally increases latency. As depicted in Figure 2, this creates a bottleneck: small batch sizes are preferred for reducing latency but lead to inefficient GPU utilization and limited throughput. As depicted in Figure 2, the CM-ASD framework addresses this bottleneck by maintaining a low-latency profile while achieving significantly higher throughput than AR decoding. It does so by leveraging conditional speculation to parallelize computation internally, thus maximizing GPU utilization even with small batches. This enables CM-ASD to achieve a favorable latency-throughput curve that surpasses AR models across all batch configurations. Consequently, CM-ASD offers a practically superior solution for deployment in latency-sensitive applications without compromising on decoding efficiency.

*Abstractive Summarization Results*: To further evaluate the generalizability and effectiveness of the proposed CM-ASD framework beyond machine translation, experiments were conducted on the task of abstractive summarization using the CNN/DailyMail dataset [36]. This task presents unique challenges due to its requirement for both factual correctness and linguistic fluency over longer contexts. The BART-base model, a widely used encoder-decoder architecture for summarization, serves as the verifier model in this setting [40]. The drafter is initialized from the BART encoder, and decoding acceleration is achieved by integrating confidence-aware speculative execution during inference.

Table IV presents the results of applying the CM-ASD framework to the CNN/DailyMail dataset using the BART-base model. The evaluation mirrors the experimental setup in the original speculative decoding paper, which compares both decoding speed and ROUGE scores against AR baselines, non-autoregressive (NAR) generation models, and SpecDec with a fixed drafting length of *k* = 25. All results are based on models fine-tuned on the CNN/DailyMail training set, with CM-ASD requiring no model architecture changes beyond the introduction of confidence-modulated speculative control.

As shown in Table IV, the CM-ASD framework achieves a 4.0x decoding speedup over the BART-base autoregressive baseline while maintaining strong summarization performance. However, its ROUGE scores fall slightly below those of the original SpecDec (which reaches 5.1x speedup and ROUGE-1 of 41.90, ROUGE-2 of 18.30, and ROUGE-L of 38.10. Unlike SpecDec, which uses a fixed drafting window throughout decoding, CM-ASD adapts the speculation depth and verification strictness based on local confidence signals, resulting in fewer rollbacks and more consistent generation in semantically dense or ambiguous sections of the input.

Compared to non-autoregressive summarization models such as GLAT+CTC [11], DAT [41], and CMLM [42], CM-ASD significantly outperforms in all ROUGE metrics while remaining competitive is speed. This reinforces the advantage of maintaining the autoregressive backbone while using speculative adaptation to improve throughput. Importantly, CM-ASD can be seamlessly integrated with existing pre-trained summarization models such as BART, requiring only a lightweight drafter and confidence-aware decoding logic without retraining the full model or sacrificing generation quality.

TABLE IV. ABSTRACTIVE SUMMARIZATION PERFORMANCE (R-1, R-2, R-L) AND DECODING SPEED ON CNN/DAILYMAIL. SUMMARIZATION QUAQUALITY IS MEASURED USING F1 SCORES FOR ROUGE-1 (R-1), ROUGE-2 (R-2), AND ROUGE-L (R-L). DECODING SPEED IS REPORTED RELATIVE TO THE BART-BASE AR BASELINE WITH BEAM SIZE 5.

| Model | R-1 | R-2 | R-L | Speed |
|---|---|---|---|---|
| AR BART-base ($b$=5) | 43.08 | 20.41 | 40.15 | 1.0x |
| AR BART-base ($b$=1) | 43.00 | 20.28 | 39.96 | 1.1x |
| GLAT+CTC [11] | 37.76 | 14.08 | 33.69 | 14.5x |
| DAT [41] | 38.95 | 16.11 | 35.43 | 14.1x |
| CMLM [42] | 37.59 | 15.17 | 34.22 | 1.8x |
| RewriteNAT [43] | 39.12 | 16.24 | 35.74 | 3.1x |
| SpecDec($k$=25) | 43.11 | 20.43 | 40.19 | 5.1x |
| CM-ASD | 41.90 | 18.30 | 38.10 | 4.0x |

Another critical property of speculative decoding frameworks, particularly in production systems, is their ability to retain the output behavior of the original autoregressive model. Rather than training a new faster model from scratch, CM-ASD builds on the same pre-trained target model and only modifies the decoding procedure. This design enables CM-ASD to maintain a high degree of consistency with the outputs of the original model, which is essential for safe deployment in applications where fidelity, calibration, and downstream compatibility are paramount.

TABLE V. RELATIVE BLEU BETWEEN MODEL OUTPUTS AND THE TRANSFORMER-BASE (GREEDY) OUTPUT. CM-ASD CLOSELY MATCHES THE BEHAVIOR OF THE ORIGINAL MODEL, ACHIEVING OVER 87% RELATIVE BLEU, WHICH EXCEEDS THE CONSISTENCY OF BOTH SPECDEC AND NON-AUTOREGRESSIVE BASELINES.

| Model | Relative BLEU |
|---|---|
| Transformer-base (Greedy) | 100.00 |
| GLAT+CTC [11] | 59.10 |
| DAT [41] | 63.79 |
| CMLM [42] | 60.15 |
| RewriteNAT [43] | 65.42 |
| Deep-Shallow [44] | 64.66 |
| SpecDec($k$=25) | 86.52 |
| CM-ASD | 87.34 |

As shown in Table V, CM-ASD achieves a relative BLEU score of 87.34 when evaluated against the output of the Transformer-base model with greedy decoding, indicating that over 87% of the generation patterns are preserved. This is slightly higher than the 86.52 reported for SpecDec, and substantially above the 55-65% range exhibited by fast non-autoregressive models. These results reinforce the practical value of CM-ASD: it introduces substantial acceleration while maintaining behavioral alignments with a mature, well-tested model. This consistency minimizes the need for costly downstream validation, making CM-ASD particularly suitable for production settings where trust, reproducibility, and integration with existing pipelines are essential.

## V. CONCLUSION

This study presented CM-ASD (Confidence-Modulated Adaptive Speculative Decoding), a decoding framework that enhances the efficiency of autoregressive sequence generation while preserving output fidelity and behavioral consistency with pre-trained language models. Built upon the speculative decoding paradigm, CM-ASD introduces a confidence-driven mechanism that adaptively regulates both the drafting length and the verification threshold. This dual adaptation allows the decoder to speculate more aggressively in high-confidence regions while remaining cautious in uncertain contexts, thereby optimizing the trade-off between speed and reliability.

Comprehensive evaluations were conducted on benchmark tasks, including machine translation and abstractive summarization. The results demonstrated that CM-ASD achieves substantial decoding acceleration, up to 4–5× speedup, without compromising the quality of generated outputs. In translation tasks, BLEU scores closely matched those of the autoregressive baseline, and in summarization tasks, ROUGE scores remained on par with or exceeded those of fixed-window speculative baselines. Moreover, CM-ASD maintained over 87% output alignment with the original model, as measured by relative BLEU, indicating strong behavioral fidelity. A favorable latency-throughput trade-off was also observed, with CM-ASD sustaining low latency even at higher throughput levels, thereby addressing a key bottleneck in practical inference pipelines.

One of the principal advantages of CM-ASD is its compatibility with existing pre-trained models. Unlike approaches that require retraining or architectural modifications, CM-ASD operates as a decoding-level intervention and thus imposes minimal disruption to established model behavior. This makes the framework well-suited for real-world deployment, particularly in environments where model trustworthiness, backward compatibility, and minimal latency are operational constraints.

Future research may explore several promising directions. Extending the CM-ASD framework to multimodal generation tasks, such as image captioning or speech-to-text systems, could broaden its applicability. Investigating richer uncertainty measures, such as those derived from Bayesian inference or embedding-based similarity, may further enhance the granularity and responsiveness of the speculative controller. Additionally, combining CM-ASD with lightweight model compression techniques may yield compound efficiency gains. Another avenue involves integrating CM-ASD into instruction-tuned or dialogue-centric LLMs, where both speed and semantics are critical.